\documentclass[letterpaper,10pt,conference]{ieeeconf}
\IEEEoverridecommandlockouts
\usepackage[T1]{fontenc}
\usepackage[utf8]{inputenc}
\usepackage{graphicx}
\usepackage{amsmath,amssymb}
\usepackage{booktabs,array}
\usepackage{siunitx}
\usepackage{tikz}
\usetikzlibrary{positioning,fit,backgrounds}
\usepackage{placeins}
\usepackage{cite}
\usepackage{xcolor}
\usepackage{soul}
\usepackage{url}
\usepackage[hidelinks]{hyperref}
\sethlcolor{yellow!35}
\graphicspath{{figures/}}

\hypersetup{
  pdftitle={Robot-Induced Bias in Movement Smoothness Metrics for Rehabilitation: Quantification and Correction in Simulation},
  pdfauthor={David Regev, Alex Scherbakov and Eran Beeri Bamani},
  pdfsubject={Submitted to the 2027 IEEE International Conference on Rehabilitation Robotics (ICORR)}
}

\title{\LARGE\bf
Robot-Induced Bias in Movement Smoothness Metrics for Rehabilitation:
Quantification and Correction in Simulation}

\author{David Regev$^{1}$, Alex Scherbakov$^{1}$ and Eran Beeri Bamani$^{1}$%
\thanks{$^{1}$Department of Information Science and Applied Artificial Intelligence,
Bar-Ilan University, Ramat Gan, Israel.\newline
Submitted to the 2027 IEEE International Conference on Rehabilitation Robotics (ICORR). This work has been submitted to the IEEE for possible publication. Copyright may be transferred without notice, after which this version may no longer be accessible.}%
}

\begin{document}
\bstctlcite{IEEEexample:BSTcontrol}
\maketitle
\thispagestyle{empty}
\pagestyle{empty}

\begin{abstract}
Robot mechanics can bias kinematic measures of recovery and limit comparisons across rehabilitation devices. We developed a closed-loop human--robot simulation calibrated to published healthy and stroke group kinematics. Paired runs held motor-control parameters and movement plans fixed to quantify bias in seven metrics relative to a low-impedance reference, normalized using external repeatability benchmarks. Within a subset informed by published device specifications, median log dimensionless jerk (LDLJ) bias was $0.93$ and $1.79$ benchmark units for stroke and healthy profiles, respectively. Sobol analysis identified Coulomb friction and apparent mass as leading contributors; LDLJ bias persisted across 11 robustness variants. LDLJ bias increased as simulated movements became less fragmented. If recovery similarly reduces fragmentation, this effect could partly mask improvement in measured smoothness. Correction combining identified robot dynamics and observed metrics with gradient-boosted trees reduced median normalized absolute error from $0.77$ to $0.10$ for LDLJ and from $0.24$ to $0.08$ for spectral arc length (SPARC) in the device-informed subset of a random 20\% hold-out. Median smoothness residuals remained at or below $0.15$ benchmark units across four separately held-out parameter regions. Under an alternative controller, LDLJ error decreased from $1.60$ to $0.17$ without retraining, although transfer varied across metrics. These simulation results support mechanical characterization and movement-dependent correction toward more comparable robotic assessments, motivating validation in patients.
\end{abstract}

\section{Introduction}
\label{sec:introduction}

\begin{figure*}[t]
  \centering
  \resizebox{\textwidth}{!}{\input{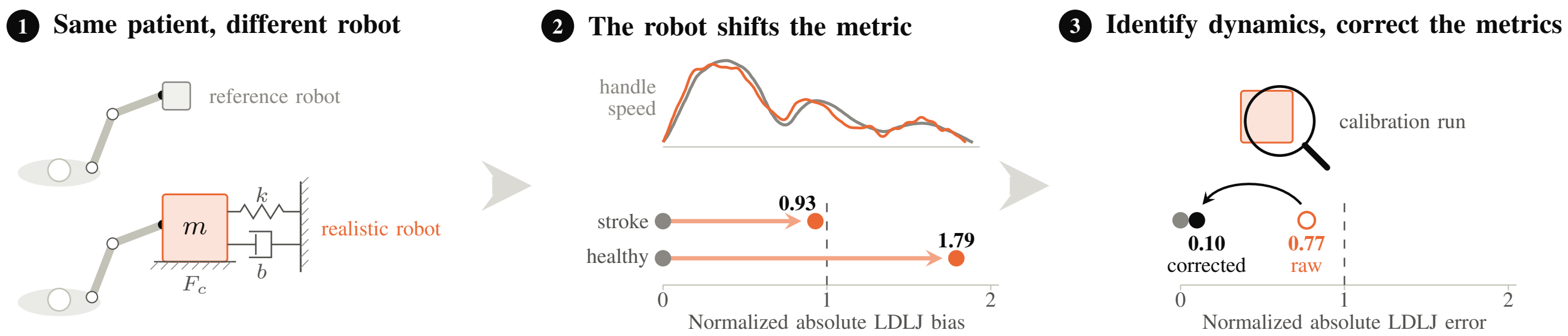}}
    \caption{Study overview. Paired simulations quantify robot-induced bias relative to a low-impedance reference. The middle panel shows simulated speed profiles and LDLJ bias. Robot identification and learned correction estimate the corresponding reference-device metrics. LDLJ values are medians in benchmark units for the device-informed subset; correction results use the random 20\% hold-out of the correction population.}
  \label{fig:overview}
  \vspace{-0.4cm}
\end{figure*}

Robotic devices are increasingly used not only to deliver therapy but also to measure recovery. End-effector robots such as MIT-MANUS record hand kinematics, yielding measures of smoothness, speed, straightness, and endpoint accuracy that track recovery after stroke and correlate with clinical scales \cite{rohrer2002smoothness,bosecker2010kinematic}. Test--retest studies provide minimal detectable change (MDC) estimates for interpreting changes within specific robotic assessment protocols \cite{gilliaux2014reaplan}. Video-based gait and balance assessment also relies on reliable clinical measures \cite{bamani2026diffusion,bamani2026bridge,stec2026reliability}; repeatability alone does not establish cross-device comparability. Interpreting unadjusted differences across devices as differences in patient function assumes that robot mechanics do not materially influence these measures. That assumption is questionable, because the robot is part of the dynamical system that produces the movement. Its apparent mass, viscous and Coulomb friction, and residual controller impedance load the arm during every reach. Such effects can introduce systematic shifts that repeated measurements on the same device may reproduce rather than reveal, making repeatability alone insufficient to establish comparability. Robot dynamics can alter wrist smoothness in ways that depend on movement configuration and metric selection \cite{rose2018assessing}. In stroke survivors, higher peak acceleration and mean jerk were observed on the low-impedance DORIS robot than on the HapticMaster \cite{oh2021doris}. A recent systematic review further highlights the need for consistent methods to evaluate robot transparency and support comparisons across studies and devices \cite{moscatelli2025transparency}.

What remains open is (i) how large these effects are across standard assessment metrics, expressed against repeatability benchmarks; (ii) which robot properties drive them and how their effects interact with motor control; (iii) how compensation for robot dynamics changes the bias; and (iv) how accurately this bias can be corrected. Simulation enables matched comparisons of the same motor-control model across parameterized robots, isolating mechanical effects while systematically varying patient and recording factors. This is particularly useful when motor-control variability and adaptation would otherwise make it difficult to attribute a metric difference to a specific mechanical property. Quantifying these effects provides a basis for interpreting device-dependent scores and identifying assessment metrics that are less sensitive to robot mechanics.

We address these questions through a closed-loop simulation of reaching with parameterized end-effector robots. Figure~\ref{fig:overview} illustrates the three stages of the approach. The first stage pairs each simulated reaching condition with a run on the low-impedance reference robot, preserving the motor-control parameters and movement plan. The second stage quantifies the resulting metric shifts; the middle panel shows simulated hand-speed profiles alongside the computed smoothness biases. Finally, the identified robot dynamics and measured movement features are used to predict and subtract the bias, yielding an estimate of the corresponding reference-device metric. Our main contributions are:
\begin{enumerate}
    \item A simulation of human-robot reaching calibrated to published healthy and stroke group kinematics, incorporating delayed feedback, motor noise, and different levels of compensation for robot dynamics.

    \item A quantitative analysis of robot-induced bias in seven assessment metrics, normalized using published MDC values and repeatability-derived benchmarks, with global Sobol sensitivity analysis to identify the contributions of robot, motor-control, and recording parameters.

    \item A correction method that combines robot dynamics identification with measured movement features to estimate metrics on a specified low-impedance reference robot. On a random 20\% hold-out of simulated cases, it reduces median absolute bias in log dimensionless jerk (LDLJ) from $0.77$ to $0.10$ benchmark units. Further evaluations compare the method with a regression baseline and test its transfer to held-out conditions and an alternative motor-control model.
\end{enumerate}

\section{Method}
\label{sec:method}

\subsection{Problem Formulation}
\label{sec:problem_formulation}

We consider planar reaching in a closed-loop human-robot simulation. The robot is parameterized by $\boldsymbol{\theta}_r=(m_x,m_y,b,F_c,k)^\top$, where $m_x$ and $m_y$ are directional apparent masses, $b$ is viscous friction, $F_c$ is Coulomb friction, and $k$ is residual centring stiffness. The motor-control parameters $\boldsymbol{\theta}_h$ govern movement planning, feedback delay, motor noise, strength, arm impedance, and compensation for robot dynamics. The internal model includes a fraction $\alpha\in[0,1]$ of the robot's dynamics, ranging from absent to complete representation. Handle trajectories emerge from the interaction of motor control and robot mechanics.

Let $\mathbf{y}(\boldsymbol{\theta}_h,\boldsymbol{\theta}_r) \in\mathbb{R}^{7}$ denote the assessment metrics computed from handle trajectories and averaged across reaching targets. A friction-free, low-impedance reference robot $\boldsymbol{\theta}_{r,0}$ has an apparent mass of \SI{0.33}{\kilo\gram}. The robot-induced bias is
\begin{equation}
    \Delta_j =
    y_j(\boldsymbol{\theta}_h,\boldsymbol{\theta}_r)
    -
    y_j(\boldsymbol{\theta}_h,\boldsymbol{\theta}_{r,0})
    \label{eq:paired_bias}
\end{equation}
for $j=1,\ldots,7$. Each pair shares the motor-control parameters, movement plan, targets, random seeds, and recording settings, isolating the effect of changing the robot under matched conditions. We quantify how this bias depends on robot mechanics, motor control, and recording conditions, and estimate the corresponding reference-device metrics. Correction uses only identified robot parameters $\widehat{\boldsymbol{\theta}}_r$ and the observed metric vector $\mathbf{y}$:
\begin{equation}
    \widehat{y}_{j,0}
    =
    y_j -
    g_j(\widehat{\boldsymbol{\theta}}_r,\mathbf{y})
    \label{eq:reference_metric_estimate}
\end{equation}
where $g_j$ predicts the signed bias of metric $j$. Paired simulations provide the reference-device values for training and evaluation.

\subsection{Reference Data and Simulation Protocol}
\label{sec:data_protocol}

\begin{figure*}[h]
\centering
\includegraphics[width=\textwidth]{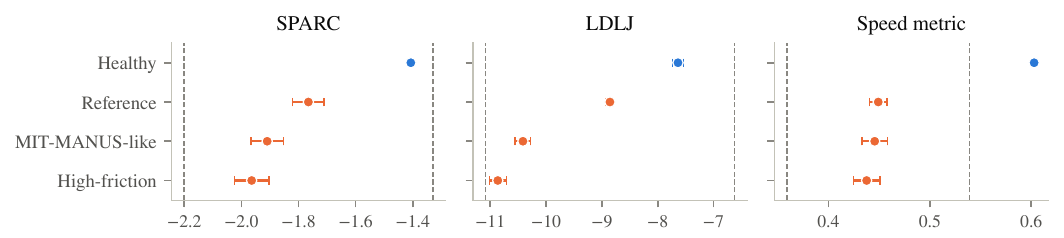}
\caption{Assessment metrics for the simulated stroke profile under adapted control on the reference, MIT-MANUS-like (0.67/1.33\,kg, 0.8\,N Coulomb friction), and high-friction (1.33\,kg, 1.11\,N Coulomb friction, 3\,N\,s/m viscous friction) robots. Healthy reference reaches provide a comparison. Markers show means with approximate 95\% confidence intervals; dashed lines indicate $\pm D_j$ around the stroke reference.}
\label{fig:headline}
\vspace{-0.4cm}
\end{figure*}

Calibration used published group means and between-subject standard deviations from the REAplan study of Gilliaux et al.\ \cite{gilliaux2014reaplan}, comprising 25 healthy participants and 25 participants with stroke. The calibration targets were peak speed, mean speed, endpoint error, straightness, and the speed metric. Test--retest MDC estimates from the same study and within-subject variability reported by Cornec et al.\ \cite{cornec2024smoothness} provided the normalization benchmarks defined in Section~\ref{sec:bias_sensitivity}. Cornec's SPARC value also provided a comparison outside the calibration objective. The simulated assessment comprised centre-out reaches to eight targets located \SI{14}{\centi\metre} from the workspace centre \cite{bosecker2010kinematic,gilliaux2014reaplan}. Handle position was resampled to \SI{200}{\hertz} and filtered using a fourth-order zero-phase Butterworth filter with a nominal \SI{10}{\hertz} cutoff. Movements were segmented at 5\% of peak speed. Metrics were computed per reach and averaged across targets, using the matched robot-reference conditions defined in Section~\ref{sec:problem_formulation}.For Fig.~\ref{fig:headline}, each condition included 48 reaches, with six repetitions per target. Each reach used a distinct random seed, with corresponding seeds shared across the four conditions. Approximate 95\% confidence intervals were computed as $\bar{y}\pm1.96\,s/\sqrt{48}$, where $\bar{y}$ and $s$ denote the mean and standard deviation across the 48 reaches.

Three simulation datasets supported the analyses: 6,144 parameter configurations per calibrated profile for global sensitivity analysis; 8,192 simulated person-robot configurations for correction development and evaluation; and 1,024 additional configurations generated with an alternative motor-control model for transfer evaluation. Parameter sampling and evaluation procedures are detailed in Sections~\ref{sec:bias_sensitivity} and~\ref{sec:metric_correction}.

\subsection{Human-Robot Model and Calibration}
\label{sec:model_calibration}

We simulated a planar two-link shoulder-elbow arm coupled to a two-degree-of-freedom end-effector robot in the horizontal plane using MuJoCo 3.15.0 \cite{todorov2012mujoco}, with a \SI{1}{\milli\second} physics step and \SI{1}{\kilo\hertz} controller. Segment lengths, masses, centre-of-mass locations, and inertias followed Winter's anthropometric tables \cite{winter2009biomechanics}, scaled to \SI{75}{\kilo\gram} and \SI{1.75}{\metre}. The grip was at mid-hand, with the workspace centre \SI{0.42}{\metre} ahead of the shoulder. A hand--handle spring-damper coupling represented grip compliance and yielded the interaction force, with stiffness \SI{3000}{\newton\per\metre} and damping \SI{30}{\newton\second\per\metre}. Healthy reaching used a minimum-jerk hand-space plan \cite{flash1985coordination}; stroke reaching used overlapping minimum-jerk submovements of decreasing amplitude, with random direction errors except on the final submovement \cite{rohrer2002smoothness}. Feedforward torques included the arm's inverse dynamics and a fraction $\alpha$ of the robot's dynamics. Intrinsic joint stiffness and viscosity acted without delay \cite{shadmehr1994adaptive}, alongside joint-space feedback with a nominal \SI{100}{\milli\second} delay. Signal-dependent motor noise \cite{harris1998signal} was low-pass filtered with a \SI{40}{\milli\second} time constant. Planning noise affected the perceived goal and movement duration, and joint-torque limits represented strength.

Calibration used the reference data in Section~\ref{sec:data_protocol}, with $\alpha=1$ on an MIT-MANUS-like robot parameterized from published specifications \cite{krebs2004manus}. The fitted vector $\boldsymbol{\phi}$ comprised plan duration, submovement count, goal and direction noise, torque limit, feedback gain, and joint damping. For each profile, grid search minimized
\begin{equation}
    \mathcal{L}(\boldsymbol{\phi})
    =
    \sum_{j\in\mathcal{J}_{\mathrm{cal}}}
    \left(
        \frac{
            \bar{y}^{\mathrm{sim}}_j(\boldsymbol{\phi})-\mu_j
        }{s_j}
    \right)^2
    \label{eq:calibration_objective}
\end{equation}
where $\mathcal{J}_{\mathrm{cal}}$ indexes the five calibration metrics, $\bar{y}^{\mathrm{sim}}_j$ is the simulated metric, and $\mu_j$ and $s_j$ are the published group mean and between-subject standard deviation \cite{gilliaux2014reaplan}. The healthy fit used a single \SI{3.0}{\second} plan; the stroke fit used three submovements over \SI{2.5}{\second}, with \SI{1.5}{\centi\metre} goal noise and \SI{0.3}{\radian} direction noise. The stroke fit retained the healthy profile's strength, feedback gain, and joint damping. As an out-of-fit check, stroke SPARC was $-2.0$, compared with $-1.82$ reported by Cornec et al.\ \cite{cornec2024smoothness}. For comparison with Oh et al.\ \cite{oh2021doris}, the stroke profile was retimed for \SI{0.20}{\metre} reaches on a light robot (\SI{0.33}{\kilo\gram}, \SI{0.09}{\newton} friction) and an admittance-controlled robot. In the latter, handle force drove a \SI{2}{\kilo\gram} virtual mass \cite{vanderlinde2003hapticmaster} with 0-\SI{5}{\newton\second\per\metre} damping, tracked by a stiff servo. Force-input delays of 0-\SI{10}{\milli\second} included the reported HapticMaster delay \cite{kuling2015delays}. Because movement times were unreported, we tested a shared plan fitted to the light robot's reported peak velocity and device-specific plans fitted separately to each robot's reported peak velocity.

Numerical checks verified analytical arm kinematics and inverse dynamics, apparent mass, viscous terminal speed, and Coulomb breakaway. Coulomb friction used MuJoCo joint friction loss with tightened constraint impedance. Refining the physics step to \SI{0.1}{\milli\second} with the controller unchanged, switching between Euler and RK4 integration, or tightening the solver changed normalized absolute bias by at most 0.01 for LDLJ and 0.04 for SPARC. An explicit smoothed Coulomb formulation agreed within 0.04 normalized units.

\subsection{Bias Quantification and Sensitivity Analysis}
\label{sec:bias_sensitivity}

The seven metrics comprised spectral arc length (SPARC), log dimensionless jerk (LDLJ) \cite{hogan2009sensitivity}, the speed metric, straightness, endpoint error, mean speed, and peak speed. SPARC used the movement-speed profile with a maximum frequency cutoff of \SI{20}{\hertz} and an amplitude threshold of 0.05 \cite{balasubramanian2012sparc,balasubramanian2015analysis}. The speed metric was mean speed divided by peak speed; straightness was straight-line distance divided by path length; and endpoint error was the final distance to the target \cite{gilliaux2014reaplan}. Implementations were checked against analytical minimum-jerk values. The paired bias in Eq.~\eqref{eq:paired_bias} was normalized as
\begin{equation}
    R_j = \frac{|\Delta_j|}{D_j},
    \qquad D_j>0
    \label{eq:normalized_bias}
\end{equation}
where $D_j$ is an external repeatability benchmark. For endpoint error, mean and peak speed, straightness, and the speed metric, we used the Target-task MDC$_{95}$ reported for 15 participants with stroke \cite{gilliaux2014reaplan}. For SPARC and LDLJ, benchmarks derived from Cornec et al.'s within-subject variability \cite{cornec2024smoothness} were

\begin{equation}
    \begin{aligned}
        D_{\mathrm{SPARC}}
        &= 0.247\,
        \left|y^{\mathrm{stroke}}_{\mathrm{SPARC},0}\right|\\[8pt]
        D_{\mathrm{LDLJ}}
        &= 0.252\,
        \left|y^{\mathrm{stroke}}_{\mathrm{LDLJ},0}\right|
    \end{aligned}
    \label{eq:smoothness_benchmarks}
\end{equation}

Here, $y^{\mathrm{stroke}}_{j,0}$ is the calibrated stroke profile's reference-device metric. The same denominators were used for both profiles. An alternative SPARC benchmark was derived from Hermle et al.'s MDC$_{90}$ of $0.03$ on the $\log_{10}$ scale \cite{hermle2025ataxia}, corresponding to an approximately 7.2\% increase in SPARC magnitude. Applied to the stroke-profile reference magnitude of $1.77$, this yielded $D_{\mathrm{SPARC,alt}}\approx0.127$, used for both profiles. We used $R_j\geq0.25$ as an analysis threshold.

Sensitivity analysis varied ten factors: apparent mass (0.33-\SI{5}{\kilo\gram}), anisotropy $m_y/m_x$ (0.5-2), viscous friction $b$ (0-\SI{10}{\newton\second\per\metre}), Coulomb friction $F_c$ (0-\SI{3}{\newton}), residual stiffness $k$ (0-\SI{10}{\newton\per\metre}), compensation $\alpha$ (0-1), motor-noise level (0.15-0.6), arm-impedance scale (0.5-2), feedback delay (80-\SI{120}{\milli\second}), and recording-filter cutoff (6-\SI{15}{\hertz}). Mass and anisotropy were sampled on log-uniform scales. For each profile, a Sobol-based design with $N=512$ base samples and $d=10$ factors yielded $N(d+2)=6144$ configurations \cite{saltelli2010variance,herman2017salib}. Each configuration was evaluated across eight targets on the tested and reference robots under the matched conditions in Section~\ref{sec:problem_formulation}. SALib estimated first-order and total-order indices for $R_j$ over the full sampled domain. Approximate 95\% confidence intervals were obtained from 100 bootstrap resamples of the base-sample rows using seed 1. Their half-widths were $1.96$ times the standard deviation of the bootstrap index estimates. First-order indices quantify individual contributions to output variance; total-order indices include interactions. Descriptive bias summaries additionally considered a subset informed by published device specifications \cite{krebs2004manus,oh2021doris,vanderlinde2003hapticmaster}: apparent mass $\leq\SI{2}{\kilo\gram}$, $F_c\leq\SI{1.2}{\newton}$, and $b\leq\SI{5}{\newton\second\per\metre}$.

\subsection{Robot Identification and Metric Correction}
\label{sec:metric_correction}

Robot dynamics were identified in a calibration run without the simulated person attached. The handle was driven by four superposed sinusoids spanning 0.4-\SI{2.9}{\hertz}, with force and position recorded at \SI{200}{\hertz}. Parameters were fitted independently for each axis by least squares:
\begin{equation}
    F \approx
    \widehat{m}\ddot{x}
    + \widehat{b}\dot{x}
    + \widehat{F}_c\operatorname{sign}(\dot{x})
    + \widehat{k}x
    \label{eq:robot_identification}
\end{equation}
where $F$ and $x$ denote force and handle displacement. Samples with $|\dot{x}|<\SI{0.01}{\metre\per\second}$ were excluded to avoid the near-sticking regime. For each metric, a gradient-boosted tree regressor predicted the paired bias from robot parameters and the seven observed metrics, using 400 boosting iterations and a learning rate of 0.05. Training parameters were randomly perturbed at scales based on simulated identification errors: approximately 6\% for mass, 10\% for friction, and \SI{0.5}{\newton\per\metre} for stiffness. The predicted bias was subtracted according to Eq.~\eqref{eq:reference_metric_estimate}. A least-squares regression baseline used robot parameters, square-root and quadratic terms, mass-friction interactions, and products of these terms with observed peak speed, mean speed, and the speed metric.

Development used the 8,192 configurations described in Section~\ref{sec:data_protocol}, spanning plan durations of 0.6-\SI{3.5}{\second}, one to four submovements, and torque limits of 2-\SI{30}{\newton\metre}, alongside the remaining motor-control and robot factors. Evaluation used a random 20\% hold-out and four separate tests excluding a specified region from training: plan durations $\geq\SI{2.5}{\second}$ or $\leq\SI{1.0}{\second}$, torque limits $\leq\SI{4}{\newton\metre}$, or compensation $\alpha\geq0.8$. Additional tests trained on one range of motor noise, arm impedance, or feedback delay and evaluated on another. Transfer was further tested on 1,024 configurations generated by an alternative optimal-feedback controller \cite{todorov2002ofc}, with the correction trained only on the primary model. The alternative controller used finite-horizon linear-quadratic regulation of a hand-space point-mass model with a \SI{40}{\milli\second} muscle filter. Terminal position, velocity, and acceleration weights were 1, 0.05, and 0.001, respectively, with effort weight $10^{-5}$. An observer corrected the forward model at \SI{8}{\per\second} using hand-state feedback delayed by 80-\SI{120}{\milli\second}. Noise, impedance, planning variability, and strength followed the primary parameterization. Independently of the clinical calibration, the controller was configured for single-peaked \SI{14}{\centi\metre} reaches without a prescribed trajectory.

Correction accuracy was measured by
\begin{equation}
    E_j =
    \frac{|\widehat{y}_{j,0}-y_{j,0}|}{D_j}
    \label{eq:correction_residual}
\end{equation}
where $\widehat{y}_{j,0}$ and $y_{j,0}$ are the estimated and paired reference-device metrics, and $D_j$ is the benchmark defined in Section~\ref{sec:bias_sensitivity}. Uncorrected measurements and both correction methods were compared using median normalized errors and proportions exceeding the 0.25 analysis threshold, within the device-informed robot-parameter subset.

\section{Results}
\label{sec:Results}

\subsection{Robot-Induced Assessment Bias}
\label{sec:results_bias}

Robot mechanics reduced measured smoothness even under adapted control. Figure~\ref{fig:headline} compares the same simulated stroke profile on the reference robot and two representative configurations: MIT-MANUS-like and high-friction. Relative to the reference, LDLJ shifted by 0.70 and 0.90 benchmark units, respectively, and SPARC by 0.33 and 0.46. The LDLJ shifts corresponded to 128\% and 165\% of the simulated healthy-stroke difference, despite unchanged motor-control parameters and movement plans. By comparison, the speed-metric shifts were only 2\% and 7\% of the corresponding healthy-stroke difference.

\begin{figure}[h]
\centering
\includegraphics[width=\columnwidth]{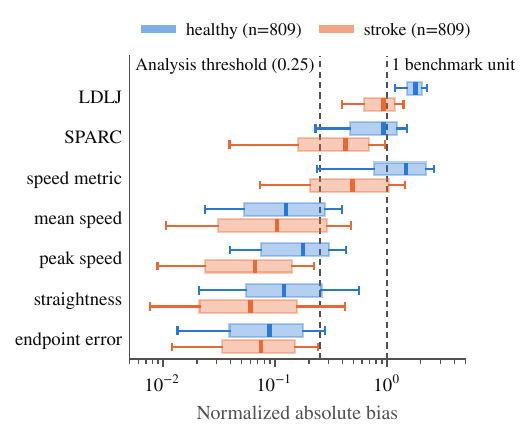}
\caption{Normalized absolute bias $R_j$ across the device-informed subset for healthy and stroke profiles. Boxes show the 25th-75th percentiles and whiskers the 10th-90th percentiles. SPARC uses the primary repeatability-derived benchmark.}
\label{fig:bias}
\vspace{-0.4cm}
\end{figure}

Figure~\ref{fig:bias} and Table~\ref{tab:mdc} summarize normalized absolute bias across the device-informed subset. Median LDLJ bias was 0.93 benchmark units for the stroke profile and 1.79 for the healthy profile; 95\% and 100\% of the sampled configurations, respectively, exceeded the 0.25 analysis threshold. Using the primary repeatability-derived benchmark, median SPARC bias was 0.43 and 0.93. Median speed-metric bias was 0.49 and 1.48; for configurations with $\alpha \geq 0.8$, the corresponding medians were 0.30 and 1.13. Each of the remaining four metrics had a median bias no greater than 0.10 for the stroke profile and 0.18 for the healthy profile.

\begin{table}[t]
\centering
\caption{Normalized absolute bias in the device-informed subset: median and 10th--90th percentiles. Benchmarks are defined in Section~\ref{sec:bias_sensitivity}.}
\label{tab:mdc}
\footnotesize
\setlength{\tabcolsep}{4pt}
\begin{tabular}{lcc}
\toprule
Metric & Stroke & Healthy \\
\midrule
SPARC, primary & 0.43 (0.04--0.97) & 0.93 (0.23--1.50) \\
SPARC, alternative & 1.46 (0.13--3.31) & 3.19 (0.79--5.14) \\
LDLJ & 0.93 (0.39--1.40) & 1.79 (1.18--2.26) \\
Speed metric & 0.49 (0.07--1.44) & 1.48 (0.23--2.62) \\
Other four (max. median) & 0.10 & 0.18 \\
\bottomrule
\end{tabular}
\vspace{-0.4cm}
\end{table}

\subsection{Sensitivity and Robustness}
\label{sec:results_sensitivity}

\begin{figure*}[t]
\centering
\includegraphics[width=\textwidth]{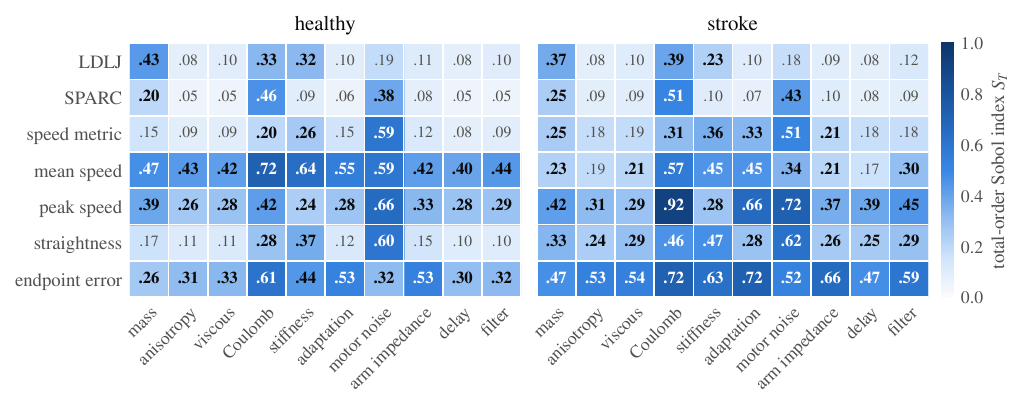}
\caption{Total-order Sobol indices $S_T$ for normalized absolute bias $R_j$ over the full sampled parameter domain, including factor interactions. Bold: $S_T\geq0.2$. Stroke peak-speed indices were not interpreted because their confidence intervals were comparable to the estimates.}
\label{fig:sobol}
\vspace{-0.4cm}
\end{figure*}

Figure~\ref{fig:sobol} identifies the principal contributors to variation in normalized absolute bias. Coulomb friction had the largest total-order index for SPARC in both profiles ($S_T=0.51\pm0.08$ for stroke, 95\% CI; $0.46$ for healthy), followed by motor noise ($0.43$ and $0.38$, respectively). For LDLJ, Coulomb friction ($0.39\pm0.06$, 95\% CI) and apparent mass ($0.37$) contributed comparably in the stroke profile. In the healthy profile, mass ranked first ($0.43$), followed by friction ($0.33$) and residual stiffness ($0.32$). Compensation for robot dynamics ranked among the leading factors for endpoint error, mean speed, and the speed metric in the stroke profile.

Controlled comparisons on 24 randomly sampled robots from the device-informed subset clarified the dependence on movement fragmentation. Adding three submovements to the healthy profile reduced LDLJ bias from 1.79 to 0.86 benchmark units; removing submovements from the stroke profile increased it from 0.91 to 1.56. The corresponding SPARC changes were 0.87 to 0.46 and 0.46 to 0.68. Across submovement counts, robot-induced dimensionless-jerk increments remained similar (within 30\%), while reference-device dimensionless jerk increased sevenfold. Consistent with LDLJ's logarithmic formulation, comparable jerk increments therefore produced smaller metric shifts against more fragmented reference movements.

LDLJ bias persisted across all 11 robustness variants covering friction formulation, recording filter, grip stiffness, arm impedance, and motor noise. For Coulomb friction of 0.56-\SI{1.11}{\newton}, under both naive and adapted control, LDLJ shifts ranged from 0.36 to 1.18 benchmark units in the stroke profile and from 1.31 to 2.13 in the healthy profile, remaining above the 0.25 analysis threshold. SPARC showed greater dependence on the tested assumptions, with shifts of 0.03-0.99 and 0.22-1.53, respectively; the stroke effect approached zero at \SI{0.56}{\newton} with doubled arm impedance or halved motor noise.

\subsection{Correction Performance}
\label{sec:results_correction}

Table~\ref{tab:corr} compares the correction methods on the random 20\% hold-out of the correction population. The proposed correction reduced median normalized absolute error from 0.77 to 0.10 for LDLJ and from 0.24 to 0.08 for SPARC, whereas the regression baseline yielded residuals of 0.26 and 0.23, respectively. The proportions exceeding the 0.25 analysis threshold fell from 74\% to 13\% for LDLJ and from 49\% to 11\% for SPARC. Median speed-metric error also decreased, from 0.58 to 0.21 benchmark units. All summaries refer to the device-informed subset. Across the four separately held-out parameter regions defined in Section~\ref{sec:metric_correction}, median residuals were 0.10-0.15 for LDLJ and 0.06-0.14 for SPARC. When motor noise, arm impedance, or feedback delay differed between training and testing, median residuals across the two smoothness metrics ranged from 0.09 to 0.22 benchmark units. Motor-noise shifts produced up to 46\% of LDLJ residuals above 0.25; training on higher noise increased SPARC error for low-noise cases from 0.12 to 0.15.

Correction trained solely on the primary motor-control model also reduced smoothness errors under the alternative optimal-feedback-control model, without retraining. Median LDLJ error decreased from 1.60 to 0.17 benchmark units, and SPARC error from 0.37 to 0.20. The proportions above the 0.25 threshold were 34\% and 39\%, respectively. Transfer depended on the metric: the speed-metric residual was 0.40, while peak-speed error increased from 0.14 to 0.23.

\begin{table}[t]
\centering
\caption{Median normalized absolute errors in the device-informed
subset for raw measurements, regression (Reg.), and
gradient-boosted trees (GBT).}
\label{tab:corr}
\footnotesize
\setlength{\tabcolsep}{3pt}
\begin{tabular}{lccc|cccc}
\toprule
 & \multicolumn{3}{c|}{Random 20\%}
 & \multicolumn{4}{c}{GBT, held-out conditions} \\
Metric & raw & reg. & GBT & slow & fast & weak & adapt. \\
\midrule
Endpoint error & 0.09 & 0.09 & 0.08 & 0.10 & 0.09 & 0.09 & 0.07 \\
Peak speed     & 0.11 & 0.10 & 0.07 & 0.07 & 0.16 & 0.06 & 0.07 \\
Mean speed     & 0.15 & 0.11 & 0.09 & 0.09 & 0.28 & 0.08 & 0.09 \\
Straightness   & 0.10 & 0.08 & 0.06 & 0.09 & 0.15 & 0.06 & 0.06 \\
Speed metric   & 0.58 & 0.41 & 0.21 & 0.42 & 0.33 & 0.21 & 0.21 \\
SPARC          & 0.24 & 0.23 & 0.08 & 0.14 & 0.06 & 0.08 & 0.07 \\
LDLJ           & 0.77 & 0.26 & 0.10 & 0.15 & 0.14 & 0.11 & 0.10 \\
\bottomrule
\end{tabular}
\vspace{-0.4cm}
\end{table}

\section{Conclusions}
\label{sec:conclusions}

Paired simulations show that robot mechanics can shift assessment scores while the motor-control profile and movement plan remain fixed. Coulomb friction and apparent mass drove smoothness bias, with LDLJ effects persisting across 11 robustness variants. Removing submovements from the simulated stroke profile increased absolute LDLJ bias from $0.91$ to $1.56$ benchmark units. If recovery similarly reduces fragmentation, an increasing downward robot-induced shift could partly mask improvement on the same device. This dependence motivates correction that accounts for movement characteristics as well as robot mechanics. Correction using identified robot dynamics and observed movement features reduced median normalized absolute error from $0.77$ to $0.10$ for LDLJ and from $0.24$ to $0.08$ for SPARC on the random hold-out of simulated cases in the device-informed subset. LDLJ improvement was retained under an alternative controller, while transfer varied across metrics. Reporting identified mass, friction, and stiffness alongside assessment scores would help interpret differences across devices. Together, these results support mechanical characterization and correction relative to a specified reference device as a route toward more comparable robotic assessments.

With a shared movement plan, the model predicted a light/heavy peak-speed ratio of 0.9-1.0, below the $1.35$ reported by Oh et al.\ \cite{oh2021doris}. Changes in planned tempo or strategy are one possible explanation; that study did not report SPARC or LDLJ. These findings concern group-calibrated planar simulations, and repeatability benchmarks provide comparison scales rather than validated clinical thresholds. Counterbalanced patient studies using identified robots or controlled friction should test whether correction reduces device-dependent differences relative to protocol-specific test-retest variability.



\IEEEtriggeratref{0}
\bibliographystyle{IEEEtran}
\bibliography{references}

\end{document}